# 27 A Review of Feature and Data Fusion with Medical Images

*Alex Pappachen James and Belur V. Dasarathy*

## CONTENTS

## 27.1 INTRODUCTION

The fusion techniques that utilize multiple feature sets to form new features that are often more robust and contain useful information for future processing are referred to as feature fusion [1]. The term data fusion is applied to the class of techniques used for combining decisions obtained from multiple feature sets to form global decisions [2]. Feature and data fusion interchangeably represent two important classes of techniques that have proved to be of practical importance in a wide range of medical imaging problems.

There has been a significant growth in the amount of scientific literature on the fusion of medical images in general since the last decade [3,4]. This largely reflects the wider importance gained in the use of medical images and multiple imaging modalities in the clinical assessment of organ conditions. In addition, the noninvasive nature of medical imaging makes it an alternative to classical techniques of drug-induced patient assessment or invasive measurement techniques. Medical images of human organs and cells from different modalities indicate different types of features and details. The use of multiple images can reveal a wide range of useful information that is not otherwise visible from a single image modality. However, going through the details in an individual modality one at a time can lead to significant time lags and requires multiple levels of expertise, making this an expensive process for the patient and the health service provider. Multimodal and multisensory imaging systems can reduce the overhead to information processing through feature and data fusion techniques to improve the overall operational efficiency.

A large variety of imaging modalities are in use today, such as magnetic resonance imaging (MRI) [5–21], computerized tomography (CT) [8,13,15,18,20,22–31], positron emission tomography (PET) [17,32–44], single-photon emission computed tomography (SPECT) [7,8,10,26,28,30,32–34,45–62], and ultrasound (US) [22,41,63–76]. Among others, they largely find applications in the study of the brain [7,11,32–35,45,46,77–99], breast [28,62,73,100–111], prostate [25,41,52,54,56,63–68,70,72,112–127], and lungs [43,44,49,128–135].

The field of medical image fusion is faced with the problems of veracity, velocity, and volume of the data that require faster and efficient processing of information. This review chapter provides an overview of information fusion techniques making use of feature and data fusion principles that find application in medical image computing and analysis. The aim of this chapter is to provide a collective view of the applicability and progress of information fusion techniques in medical imaging useful for clinical studies [3,4,22,136–144].

## 27.2   FEATURE-LEVEL MEDICAL IMAGE FUSION  METHODS

We organize the methodological developments in medical image fusion methods into those that rely on feature-level processing and those that work at decision-level fusion. Feature-level fusion often aids in improving image quality and extracts newer features that are otherwise difficult to find in the original set of features.

Feature-level fusion between images is challenged by the problem of interimage variability such as pixel mismatches (scale, rotations, shifts), missing pixels, image noise, resolution, and contrast. The inaccuracies in feature representations can lead to poor fusion performance and lesser robustness of the feature representation. In addition, this also means that wrong feature representations can lead to wrong conclusions (increased false positives and false negatives) that reduce the reliability of medical image analysis in clinical settings.

Figure 27.1 is a summary of the major medical image fusion methods that are used individually and in combination for solving clinically relevant medical imaging computing problems.

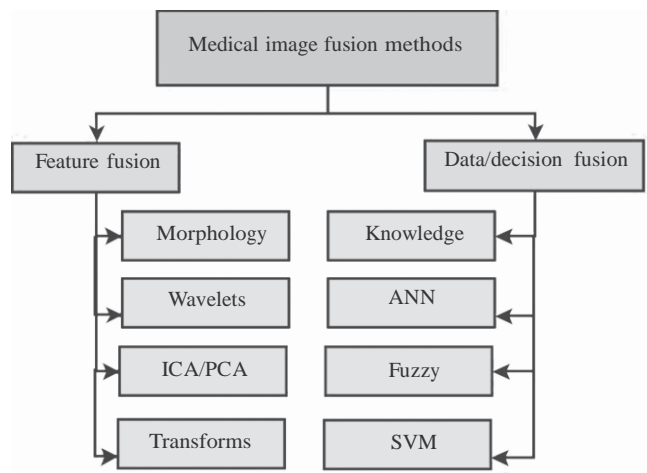

**FIGURE 27.1**   The classification tree for the major list of medical image fusion methods.

### 27.2.1 Morphological Operators and Filters

Morphological operators make use of the connectedness between pixels either to improve the spatial arrange of the pixels or to distort them to extract useful features from the subset of spatially localized pixel features. The filters designed with morphological operators have been successfully applied in the problem of diagnosis of brain conditions to analyze and identify tumors [32,77,91]. The morphological operators are used for fusing the images from multiple modalities such as CT and MR [77,78], with a varied degree of success. The success of these operators depends on the size and design of the structuring operator that invariably controls the opening and closing operations in morphological filtering. Among many, the major operators used for fusion are averaging, morphology towers, K-L transforms, and morphology pyramids. The advantage of the morphological operators results from their simplicity and ability to parallelize for high-speed implementations, while the drawbacks largely result from the high dependence of pixel intensities.

### 27.2.2 Wavelet-Based Feature Fusion

Wavelet transforms have the ability to compress the details of the images through their coefficients and to separate the fine and coarse details from one another. Because of the ability to represent the different properties of the image through coefficients, the impact of noise on the image would be reflected in one coefficient or another. This opens up the possibility to use wavelets to retain only those coefficients that are stable. Such coefficients from different features can be fused together to form more robust representations of the images [5,8,13,27,29,31,145–171]. In effect, the idea of the wavelet fusion is to inject good features from one image to another and in the process remove the problematic ones. Substitution, addition, aggregator functions, and data-driven models all form the methodological process of injection. Although the coefficients do show a compressed processing, the fused output image is optimized for maximum resolution and target quality. The high resolution of the input images can lead to increased computational complexity, whereas a combination of high- and low-resolution images for fusion can make the problem of feature level fusion challenging in terms of robustness. Examples of the application of wavelets include image pseudo coloring [85], improving the resolution of the images through super-resolution techniques [8], diagnosis with medical images [27,145,152,172], lifting schemes [173], image segmentation [146], planning for radiotherapy treatment using 3-D conformal mapping [154], and color visualization for labeling [167].

### 27.2.3 Wavelet-Based Hybrid Feature Fusion

The features obtained from the wavelet feature fusion techniques have been used along with other feature extraction methods to improve the robustness of the wavelet-based fusion approaches. Neural networks, considered an excellent candidate for dimensionality reduction and feature extraction, have been employed along with fusion rules set by wavelet operators to implement medical image fusion [145,151,172]. Several combinations of the operators exist that have been combined along with wavelet operators to improve the robustness of the features. Some examples are combinations with support vector machines [150], the use of wavelet-texture measure [27], wavelet combined with magnetic resonance angiogram (MRA) [152,153], the use of wavelet-self adaptive operator [155], wavelet-resolution with entropy [156,158], nonlinear approach with properties of wavelet-shift invariant imaging [157], independent component analysis (ICA) combined with wavelet [174], wavelet and edge features [161], wavelet with a genetic approach [162], wavelet combined with contourlet transform [168], hybrid of neuron networks with fuzzy logic and wavelets [169], and wavelet entropy [171].

### 27.2.4 Component Analysis Techniques

Several dimensionality reduction methods exist that can reduce the large feature set to a smaller subset of algebraically transformed features. The idea of extracting components from the images has been widely explored via ICA [97,174] and principal component analysis (PCA) [175–178]. Like wavelet coefficients, the derived feature coefficients from these techniques can be used to reconstruct the image with only a small number of feature coefficients. They find application in higher resolution and large volume imaging such as volumetric medical imagery [179]. A multimodal image fusion based on PCA using the intensity-hue-saturation (IHS) transform has been shown to preserve spatial features and required functional information without color distortion [178].

### 27.2.5 Transform-Based Approaches

There are different mathematical transforms on features that can enhance the performance of the image fusion. For example, the combination of complex contourlet transform with wavelet has been shown to result in robust image fusion [176,177]. Transform-based methods are also applied for liver diagnosis [50], risk factor fusion [180], prediction of multifactorial diseases [180], parametric classification [180], local image analysis [181], and multimodality image fusion [168,176,177,182–184]. Possibilistic clustering methods show improvement over the fuzzy c-means clustering and have a wide range of application in registration stages of image fusion. Some of the applications of possibilistic clustering include tissue classification [87], diagnosis of brain conditions [34,185], and automatic segmentation [88].

## 27.3 DATA FUSION METHODS IN MEDICAL IMAGING

### 27.3.1 Knowledge in Data Fusion Methods

Even more often, it becomes quite a difficult premise to replace the expertise of the medical practitioner in improving and validating the computer-aided analysis of the medical images in segmentation of the regions of interest, labeling and updating the points of interest, and re-registering the images. A high level of domain-specific knowledge is required to specify the type of image and region of interest, which leads to a range of practical applications in the image analysis concerned with region segmentation [79], microcalcification diagnosis [186], classification of tissues [89], diagnosis of brain-related condition [89], classifiers for fusion [107], breast cancer and tumor detection [107], and delineation and recognition of anatomical parts of the brain [79].

### 27.3.2 Data Fusion With Artificial Neural Networks

Artificial neural networks (ANNs) represent a set of decision processing models inspired from the working of the human neural network. The neural networks consist of a weighted addition of inputs followed up with decisions at each of its nodes and further layers of neuron nodes acting as decision aggregates to global decisions. Because each node processes information from the group of input pixels, the network can learn and make decisions in modular levels. This makes it useful for a wide range of decision fusion applications that involve feature generation and classification [187], generic data fusion [145,186,187], various applications specific to image fusion [103,145,151,188–192], identification and diagnosis of microcalcification [186], breast cancer detection [103,109,193], data-driven medical diagnosis [145,172,191], cancer diagnosis [194], natural computing methods [195], and classifier fusion [193].

### 27.3.3 Data Fusion With Hybrid Artificial Neural Networks

The combination of ANNs with other fusion techniques results in hybrid-ANN methods. They usually combine feature-level decisions and fusion-level decisions with the neural network training algorithms to improve image fusion performances. The major group of techniques includes wavelets combined with the neural network [145,151,172], neural networks combined with fuzzy logic [190,192], combinations of fuzzy logic with a genetic-neural network [195], and support vector machines (SVMs) combined with ANN and Gaussian mixture model (GMM) [193].

## 27.3.4  Data Fusion With Fuzzy Logic

The fuzzy approach to decision making allows for a greater level of flexibility in the grouping of features and decisions utilizing a wide set of fuzzy set operators and membership functions for image-based decision fusion algorithms [10,18,32,34,38,88,91,93,95,162,169,190,192,195–204]. They find applications in diagnosis of brain conditions [32,34,91,196], treatment of cancer [38], image integration and segmentation [38,88], maximization of mutual information [10], deep brain stimulation [93], segmentation of brain tumors [95], feature fusion and image retrieval [197,198], weighted entropy calculations in images [197], multimodal analysis and image fusion [162,190,199], ovarian cancer detection and diagnosis [200], sensor-oriented image fusion [201], natural computing methods [195], and gene expression [202,203].

## 27.3.5  Data Fusion With Hybrid Fuzzy Logic

The optimal selection of feature sets, membership functions, and fuzzy operators remains an open problem. Similar to other hybrid approaches, fuzzy decisions can be combined with other fusion approaches to obtain hybrid-fuzzy fusion algorithms. Common examples of hybrid-fuzzy fusion methods are fuzzy-neural network [190,192], fuzzy logic combined with genetic-neural network-rough set [195], fuzzy logic with statistical probability measures [202], and fuzzy logic combined with neural networks and wavelets [169].

## 27.3.6  SVM Classifier-Based Approaches

Decision fusion is a straightforward operation when it comes to the majority of classifiers, as they inherently need to make local and global decisions to classify patterns. Most classifiers rely on thresholds to make a decision, whereas others go with statistical approaches; nonetheless ranking the scores and selecting the most likely one forms the core idea of asserting the presence or absence of a pattern. SVMs are a parameter-driven approach of detecting feature closeness and removing outliers for determining the class of the patterns. The ability to make decisions at local levels in the images is used in the process of decision fusion. Some of the applications of SVMs as a tool for image fusion include cancer diagnosis [194,205], classifier fusion [108,193,205], breast cancer diagnosis and treatment [108,193], image fusion [150,206], content-based image retrieval [207,208], tumor segmentation [206], gene classification [209], and feature fusion [208].

## 27.3.7  Hybrid SVM Classifier-Based Approaches

The SVMs can be combined along with other fusion algorithms and techniques to improve processing speed and to work with better representations of low-dimensional feature vectors. These hybrid SVM methods include SVM combined with wavelets [150], SVM with adaptive similarity measures [207], SVM-data fusion [206], and SVM combined with ANN and GMM [193].

## 27.4  DISCUSSION AND CONCLUSIONS

Image fusion studies with medical images face several challenges having a significant impact in the field of medical diagnostics and monitoring. The wide-range use of information and communication technologies in the health sciences during the last decade has increased trust in technology for image analysis as an essential tool. However, there is hardly any imaging modality that can capture all the possible mechanisms required to reveal the conditions under study. This necessitates the use of multimodal imaging techniques; however, they are limited by the significant footprint it takes on computational and human resources to improve the efficiency of decision processing and clinical conclusions.

The technological challenges with image fusion are manifold, including sensor-level errors, imaging noise, interimage variabilities, motion artifacts, contrast variations, and interimage resolution mismatches. Many of these issues also make the automated co-registration and normalization process between the images a difficult problem to solve. They become even more serious issues in real-time imaging systems where high-speed sampling along with increased imaging accuracy is essential to ensure accuracy and reliability of image fusion methods.

Feature-level fusion methods are affected by the imaging quality and the natural variability of the modality. The importance of improving the image formation methods necessitates careful attention when designing feature fusion techniques, as they constitute the primary reason for the robustness of fusion techniques across a wide range of imaging conditions. Noise estimation is another important area that is growing in significance to improve the signal quality before fusion techniques can be applied. The processing speed of the large-volume feature fusion algorithms can be improved by the practical realization of algorithms in field-programmable gate array (FPGA) or graphical processing units. They could in the future find practical applications in real-time monitoring, telemedical diagnosis, and surgery.

The decision-level fusion methods require a good set of features at most times to ensure high reliability of fusion. Because they are highly dependent on the underlying data structure, they are generally referred to as a data fusion technique. The computational complexity of a majority of decision fusion techniques increases nonlinearly with any linear increase in feature size. In the future, this can be a serious challenge, as the convenience of data-driven processing requires a large volume of images for processing to ensure high accuracy and reliability. Although the data-driven techniques can lead to robust fusion rules, the trust of the users plays a major role in the adoption of data-driven techniques in mainstream health systems.

Overall, both feature and data fusion techniques have made promising progress in the practical domains of medical diagnosis and analysis. This is evident from the large number of algorithmic and medical studies that make use of automated medical image fusion techniques. Future progress could very well depend on developing techniques that are well tested across realistic case studies and scenarios across a large collection of data. This also requires a large-scale standardization of data sets to compare techniques that can be considered reliable to be used in clinical settings. The major methods that have been shown to be useful for feature and data fusion include wavelet transforms, neural networks, ICA/PCA, fuzzy logic, morphology methods, SVMs, and their combinations. Further major progress that is required is the miniaturization of medical devices with increased processing capability and reliability. The ability of these devices to make use of modern communication technologies also plays a major role in the sustainable use of the fusion algorithms.